# Adaptation of Mamdani Fuzzy Inference System Using Neuro - Genetic Approach for Tactical Air Combat Decision Support System

Cong Tran[1], Lakhmi Jain[1], Ajith Abraham[2]

[1] School of Electrical and Information Engineering
University of South Australia, Adelaide, Australia
`tramcm001@students.unisa.edu.au, lakhmi.jain@unisa.edu.au`
[2] Faculty of Information Technology, School of Business Systems,
Monash University (Clayton Campus), Victoria 3168, Australia
`ajith.abraham@ieee.org, http://ajith.softcomputing.net`

**Abstract.** Normally a decision support system is build to solve problem where multi-criteria decisions are involved. The knowledge base is the vital part of the decision support containing the information or data that is used in decision-making process. This is the field where engineers and scientists have applied several intelligent techniques and heuristics to obtain optimal decisions from imprecise information. In this paper, we present a hybrid neuro-genetic learning approach for the adaptation a Mamdani fuzzy inference system for the Tactical Air Combat Decision Support System (TACDSS). Some simulation results demonstrating the difference of the learning techniques and are also provided.

## 1. Introduction

Several decision support systems have been applied mostly in the fields of medical diagnosis, business management, control system, command and control of defence and air traffic control and so on. For most Decision Support Systems (DSS), people normally make use of their experience or expertise knowledge. The problem becomes very interesting when no prior knowledge is available. Recently researchers have started using expert systems, fuzzy logic, rough sets and neural network learning methods to develop DSS. For a detailed review of different techniques, please refer to [2]. Our approach is based on the development of different adaptive fuzzy inference systems using several learning techniques [2] [3]. Fuzzy logic provides a computational framework to capture the uncertainties associated with human cognitive process such as thinking and reasoning [8]. The disadvantage of fuzzy inference system is the requirement of expert knowledge to set up the fuzzy rules, membership function parameters, fuzzy operators etc. Neural network learning methods [1] and evolutionary computation [2] could be used to adapt the fuzzy inference system.

In Section 2, we present the complexity of the tactical air combat environment problem followed by the modeling of the TACDSS using an adaptive Mamdani fuzzy inference system in Section 3. Section 4 deals with experimentation setup and results and some conclusions are also provided towards the end.

## 2. Decision Making in Tactical Air Combat

The air operation division of Defence Science and Technology Organisation (DSTO) and our research team has a collaborative project to develop a tactical environment decision support system for a pilot or mission commander in tactical air combat. In Figure 1 a typical scenario of air combat tactical environment is presented. The Airborne Early Warning and Control (AEW&C) is performing surveillance in a particular area of operation. It has two hornets (F/A-18s) under its control at the ground base as shown "+" in the left corner of Figure 1. An air-to-air fuel tanker (KB707) "☐" is on station and the location and status are known to the AEW&C. Two of the hornets is on patrol in the area of combat air patrol (CAP). Sometime later, the AEW&C onboard sensors detect a 4 hostile aircraft (Mig-29) that is shown as "O". When the hostile aircrafts enter the surveillance region (shown as dashed circle) the mission system software is able to identify the enemy aircraft and its distance from the Hornets in the ground base or in the CAP.

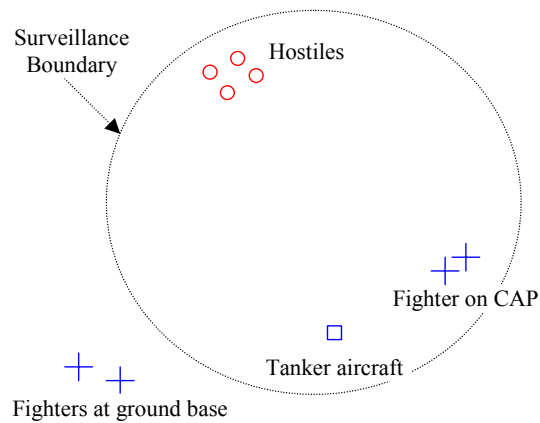

**Figure 1.** A simple scenario of the air combat

The mission operator has few options to make a decision on the allocation of hornets to intercept the enemy aircraft.

- Send the Hornet directly to the spotted area and intercept;
- Call the Hornet in the area back to ground base and send another Hornet from the ground base
- Call the Hornet in the area for refuel before intercepting the enemy aircraft

The mission operator will base his decisions on a number of decision factors, such as:
- Fuel used and weapon status of hornet in the area
- Interrupt time of Hornet in the ground base in the Hornet at the CAP to stop the hostile.

- The speed of the enemy fighter aircraft and the type of weapons it possesses.
- The information of enemy aircraft with type of aircraft, weapon, number of aircraft.

From the above simple scenario, it is evident that there are several important decision factors of the tactical environment that might directly affect the air combat decision. We simulated a tactical air combat situation and based on some expert knowledge we made use of the fuzzy neural network framework [5] to develop the TACDSS. In the simple tactical air combat, the four decision factors that could affect the decision options of hornet in the CAP or the hornet at the ground base are the following:

"*fuel status*"- quantity of fuel available to perform the intercept,
"*weapon possession status*" - state of weapon of the hornet,
"*interrupt time*"- time required that the hornet will fly to interrupt the hostile and
"*danger situation*" - information of the hornet and the hostile in the battlefield.

Each factors has difference range of unit such as the *fuel status* (0 to 1000 litres), *interrupt time* (0 to 60 minutes), *weapon status* (0 to 100 %) and *danger situation* (0 to 10 points). We used the following two expert rules for developing the fuzzy inference system.

- The decision selection will have small value if the *fuel status* is too low, the *interrupt time* is too long, the hornet has low *weapon status*, and the *danger situation* is high value.
- The decision selection will have high value if the *fuel status* is full, the *interrupt time* is fast enough, the hornet has high *weapon status* and the *danger situation* is low value.

In the air combat, decision-making is always based on all states of decision factors. But sometime, a mission operator or commander can make a decision basing on an important factor, such as the fuel used of the hornet is too low, the enemy has more power weapon, quality and quantity of enemy aircraft. Table 1 shows the score (decision selection point) at each stage of each tactical air combat decision factors.

## 3. Modeling TACDSS Using Adaptive Fuzzy Inference System

We made use of the Fuzzy Neural Network (FuNN) framework [5] for learning the Mamdani inference method. A functional block diagram of the FuNN model is depicted in Figure 2 and it consists of two phases of learning processes. The first phase is the structure-learning (*if-then* rules) phase using the knowledge acquisition module. The second phase is the parameter-learning phase for tuning membership functions to achieve a desired level of performance. FuNN uses a gradient descent-learning algorithm to fine-tune the parameters of the fuzzy membership functions. In the connectionist structure, the input and output nodes represent the input states and output control-decision signals, respectively, and in the hidden layers, there are nodes functioning as quantification of membership functions (MFs) and *if-then* rules. Please refer to [5] for details regarding architecture and function of the various layers.

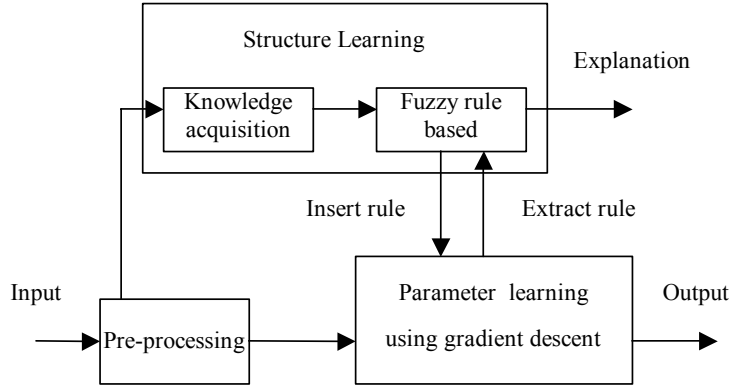

**Figure 2.** A general schematic diagram of the hybrid fuzzy neural network

We used a simple and straightforward method proposed by Wang and Mendel [6] for generating fuzzy rules from numerical input-output training data. The task here is to generate a set of fuzzy rules from the desired input-output pairs and then use these fuzzy rules to determine the complete structure of the TACDSS.

Suppose we are given the following set of desired input -($x_1$,$x_2$) output ($y$) data pairs ($x_1$,$x_2$,$y$): (0.6, 0.2; 0.2), (0.4, 0.3; 0.4). In TACDSS, input variable *fuel used* has a degree of 0.8 in *half*, a degree of 0.2 in *full*. Similarly, input variable *time intercept* has degree of 0.6 in *empty* and of 0.3 in *normal*. Secondly, assign $x_1^i$, $x_2^i$, and $y^i$ to a region that has maximum degree. Finally, obtain one rule from one pair of desired input-output data, for example,

($x_1^1$,$x_2^1$,$y^1$) => [$x_1^1$(0.8 in *half*), $x_2^1$(0.2 in *fast*),$y^1$ (0.6 in *acceptable*)],
• $R_1$: if $x_1$ is *half* and $x_2$ is *fast*, then $y$ is *acceptable*;

($x_1^2$,$x_2^2$,$y^2$), => [$x_1$(0.8 in *half*),$x_2$(0.6 in *normal*),$y^2$(0.8 in *acceptable*)],
• $R_2$: if $x_1$ is *half* and $x_2$ is *normal*, then $y$ is *acceptable*.

Assign a degree to each rule. To resolve a possible conflict problem, i.e. rules having the same antecedent but a different consequent, and to reduce the number of rules, we assign a degree to each rule generated from data pairs and accept only the rule from a conflict group that has a maximum degree. In other words, this step is performed to delete redundant rules, and therefore obtain a concise fuzzy rule base. The following product strategy is used to assign a degree to each rule. The degree of the rule denoted by

$Ri$ : if $x_1$ is $A$ and $x_2$ is $B$, then $y$ is $C(w_i)$,

The rule weight is defined as

$w_i = \mu_A(x_1)\mu_B(x_2)\mu_c(y)$

For example of TACS, $R_1$ has a degree of

$W_1 = \mu_{half}(x_1)\, \mu_{fast}(x_2)\, \mu_{acceptable}(y) = 0.8 \times 0.2 \times 0.6 = 0.096$,

and $R_2$ has a degree of

$W2 = \mu_{half}(x_1)\, \mu_{normal}(x_2)\, \mu_{acceptable}(y) = 0.8 \times 0.6 \times 0.8 = 0.384$

Note, that if two or more generated fuzzy rules have the same preconditions and consequents, then the rule that has maximum degree is used. In this way, assigning the degree to each rule, the fuzzy rule base can be adapted or updated by the relative weighting strategy: the more task related the rule becomes, the more weight degree the rule gains. As a result, not only is the conflict problem resolved, but also the number of rules is reduced significantly. After the structure-learning phase (*if-then* rules), the whole network structure is established, and the network enters the second learning phase to optimally adjust the parameters of the membership functions using a gradient descent learning algorithm to minimise the error function

$$E = \frac{1}{2} \sum_x \sum_{l=1}^{q} (d_l - y_l)^2$$

where $d$ and $y$, are the target and actual outputs for an input $x$. We also explored Genetic algorithms (GA) to optimize the membership function parameters. Given that the optimisation of fuzzy membership functions may involve many changes to many different functions, and that a change to one function may effect others, the large possible solution space for this problem is a natural candidate for a GA based approach. The GA module for adapting the membership function parameters acts as a stand-alone system that already have the if-then rules. GA optimises the antecedent and consequent membership functions. The GA used in the system is, in essence, the same as simple genetic algorithm, with the important exception that the chromosomes are represented as strings of floating point numbers, rather than strings of bits. Figure 3 depicts the chromosome architecture representing the centre of input and output MFs. One point crossover was used for the reproduction of chromosomes.

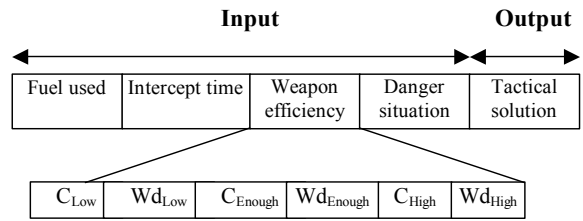

**Figure 3.** The chromosome of the centres of inputs and output MFs of TACDSS

## 4. Experimentations Results

The fuzzy inference system was created using the FuNN framework [7]. The TACDSS has four inputs and one output variable. We used triangular membership functions and each input variable were assigned three MFs. 16 fuzzy rules were created using the methodology mentioned in Section 3. With the momentum at 0.8 (after a trail and error approach), we varied the learning rates to evaluate the performance. For 10 epochs, we obtained a RMSE of 0.5775 (learning rate 0.1) and RMSE of 0.2889 (learning rate 0.3) respectively. Figure 6 shows the 16 fuzzy *if-then* rules of the developed TACDSS.

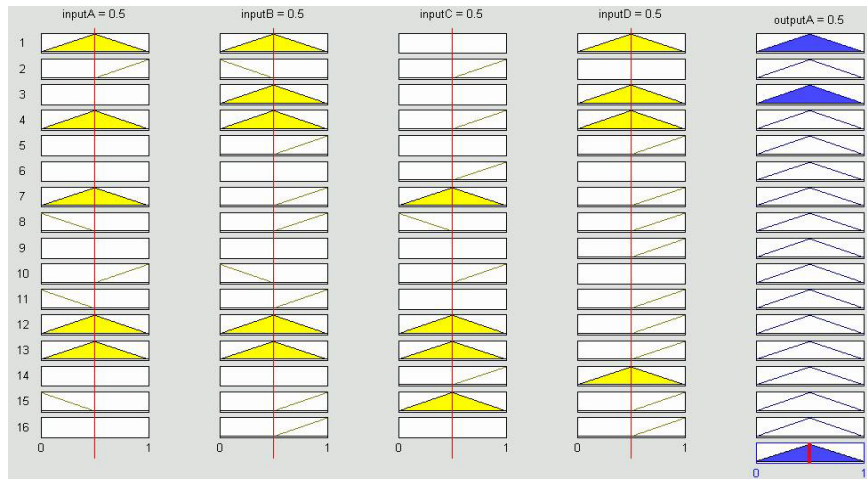

**Figure 4.** Developed Mamdani fuzzy inference system for TACDSS

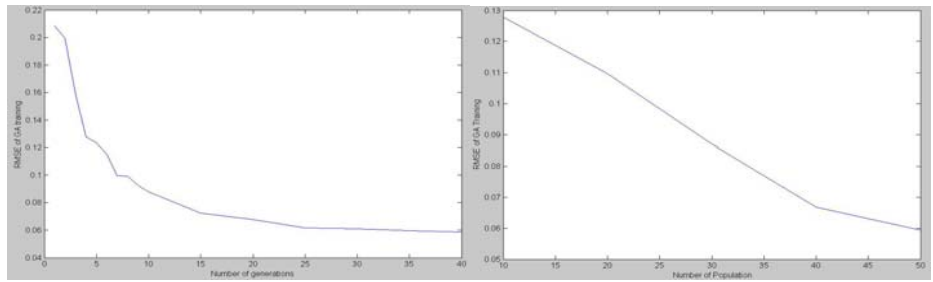

**Figure 5.** RMSE performance for number of generations and population size

We also explored the fine-tuning of membership functions using an evolutionary algorithm. We started with a population size 10, tournament selection strategy, mutation rate 0.01 and implemented a one point crossover operator. After a trial and error approach (please refer to Figure 5) by increasing the population size and the number of iterations (generations), we finalized the population size and number of iterations as 50. To improve the accuracy we extracted 49 fuzzy *if-then* rules to describe the TACDSS. We obtained an RMSE of 0.05934 after 50 generations of evolutionary learning with a population size of 50. Figure 6 demonstrates the effect of parameter tuning of membership functions (before and after evolutionary learning) for the input variable *fuel used*.

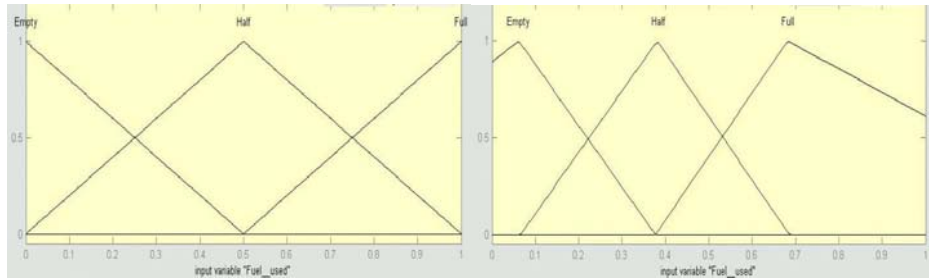

**Figure 6**. The MFs of input variable *fuel used* before and after GA learning

The developed TACDSS system for a simulated bad and good situation as depicted as follows.

**Bad situation:** When the *fuel used* is 0.05, *time intercept* is 0.95, *weapon status* is 0.05 and *danger situation* is 0.95 then *tactical decision score* is 0.416

**Good situation:** When the *fuel used* is 0.95, *time intercept* is 0.05, *weapon status* is 0.95 and *dangerous situation* is 0.05 then *tactical decision score* is 0.503

For the test set *fuel used* 0.938, *time intercept* 0.05167, *weapon status* 0.975, *danger situation* 0.124 the expected *tactical decision* is 0.939 and the developed TACDSS decision score was 0.498 (approximately 47% less than the required value).

## 5. Conclusion and future research

In this paper, we have explained a hybrid fuzzy inference model for developing a tactical air combat decision support system. Our case study of the simple scenario of the air tactical environment demonstrates the difficulties to implement the human decision-making. We have explored two learning techniques using backpropagation and evolutionary learning to fine-tune the membership function parameters. Empirical results reveal that evolutionary approach performed better in terms of low RMSE with a trade off in computational cost.

This work was an extension of our previous research wherein we have used a Takagi-Sugeno fuzzy inference model for developing the TACDSS. Empirical results on test data indicate that the Takagi-Sugeno version of TACDSS performed better than the current Mamdani version. Our future work includes development of decision trees and adaptive reinforcement learning systems that can update the knowledge base from data when no expert knowledge is available.